%% file: main.tex
\documentclass[dvipsnamesm, acmsmall, nonacm]{acmart}

\usepackage{graphicx}
\usepackage{xcolor}
\usepackage[normalem]{ulem}
\usepackage{microtype}
\usepackage{listings}
\usepackage{algorithm}
\usepackage[noend]{algpseudocode}
\usepackage{comment}
\usepackage{paralist}
\usepackage{subcaption}
\usepackage{soul}
\usepackage[T1]{fontenc}
\usepackage{dblfloatfix}
\usepackage{textcomp}
\usepackage{caption}
\usepackage{wrapfig}

\makeatletter
\let\@authorsaddresses\@empty
\makeatother

\newcommand{\hlnew}[1]{#1}%

\begin{document}
\setlength\intextsep{3pt}

%
\title{GEVO: GPU Code Optimization using Evolutionary Computation}

\author{Jhe-Yu Liou}
\affiliation{%
   \institution{Arizona State University, jhe-yu.liou@asu.edu}
   \city{Tempe}
   \state{AZ}}
\email{jhe-yu.liou@asu.edu}
\author{Xiaodong Wang}
\affiliation{%
   \institution{Facebook}
   \city{Menlo Park}
   \state{CA}}
\author{Stephanie Forrest}
\affiliation{%
   \institution{Arizona State University}
   \city{Tempe}
   \state{AZ}}
\affiliation{%
   \institution{Santa Fe Institute}
   \city{Santa Fe}
   \state{NM}}
\author{Carole-Jean Wu}
\affiliation{%
   \institution{Arizona State University}
   \city{Tempe}
   \state{AZ}}
\affiliation{%
   \institution{Facebook}
   \city{Menlo Park}
   \state{CA}}

\input{abstract}



\maketitle

\input{introduction}
\input{related}
\input{design.tex}
\input{experimental.tex}
\input{result}
\input{discussion}

\begin{acks}

We thank F. Esponda, W. Weimer, 
E. Schulte, and the reviewers for many insights, code and helpful comments.
This work is supported in part by the National Science Foundation under CCF-1618039 SHF-1652132, NSF CCF 1908633; DARPA FA8750-19C-0003, and
AFRL FA8750-19-1-0501.
\end{acks}

\bibliographystyle{ACM-Reference-Format-abbrev}
\bibliography{ref}

\end{document}

%% file: abstract.tex
\begin{abstract}
GPUs are a key enabler of the revolution in machine learning and high performance computing, functioning as de facto co-processors to accelerate large-scale computation. 
As the programming stack and tool support have matured, GPUs have also become accessible to programmers, who may lack
detailed knowledge of the underlying architecture and fail to
fully leverage the GPU's computation power. 
GEVO (Gpu optimization using EVOlutionary computation) is a tool for
automatically discovering optimization opportunities and tuning the performance of GPU kernels in the LLVM representation. 
GEVO uses population-based search to find edits to GPU code compiled to LLVM-IR and improves performance on desired criteria while retaining required functionality. 
We demonstrate that GEVO improves the execution time of the GPU programs in the Rodinia benchmark suite and the machine learning models, SVM and ResNet18, on NVIDIA Tesla P100. For the Rodinia benchmarks, GEVO improves GPU kernel runtime performance by an average of 49.48\% and by as much as 412\% over the fully compiler-optimized baseline. If kernel output accuracy is relaxed to tolerate up to 1\% error, GEVO can find kernel variants that outperform the baseline version by an average of 51.08\%. 
For the machine learning workloads, GEVO achieves kernel performance improvement for SVM on the MNIST handwriting recognition (3.24X) and the a9a income prediction (2.93X) datasets with no loss of model accuracy. GEVO achieves 1.79X kernel performance improvement on image classification using ResNet18/CIFAR-10, with less than 1\% model accuracy reduction. 

\end{abstract}

%% file: introduction.tex
\section{Introduction}
\label{sec:introduction}
\vspace{-0.1cm}

\begin{wrapfigure}{R}{0.5\textwidth}
  \includegraphics[width=0.48\textwidth]{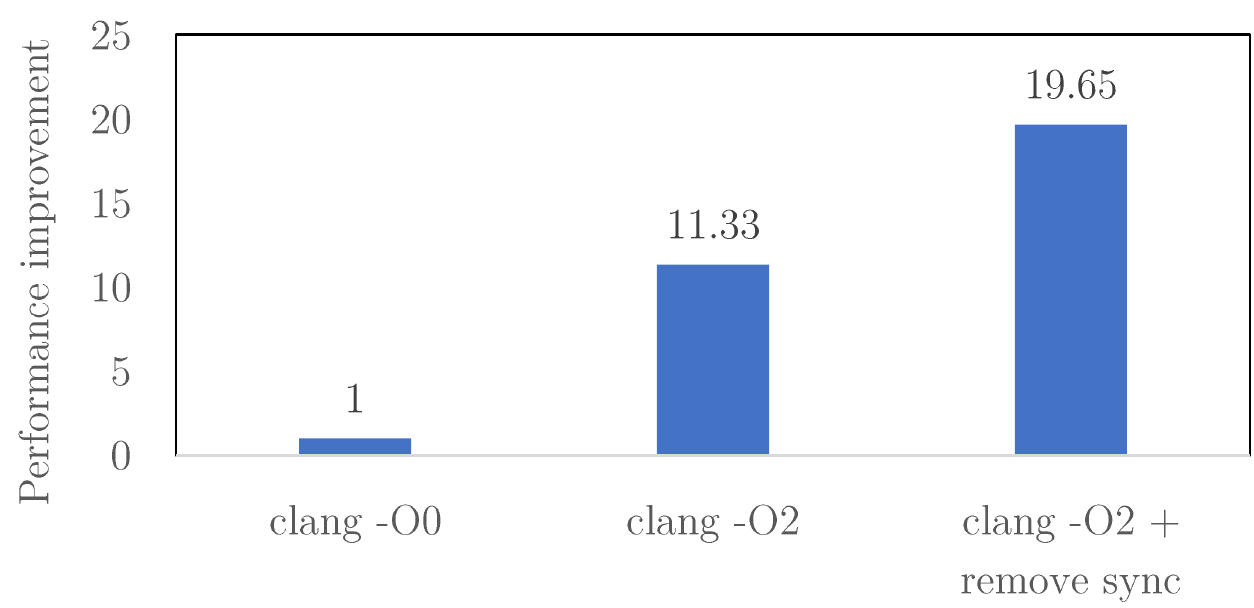}
  \vspace{-0.3cm}
  \caption{
    \hlnew{Performance improvements for {\it nw} with no compiler optimization (clang -O0), with full compiler optimization (clang -O2), and when unneeded synchronization primitives are removed manually.}}
    \label{fig:nw}
\end{wrapfigure}

The fields of large-scale machine learning and scientific algorithms are expanding quickly and pushing the limits of high performance computing. Continued advances in these fields will require orders-of-magnitude improvements in computing power~\cite{dhillon2002data,judd1998large,wang2008scientific}.
GPUs help address this challenge and have become de-facto co-processors~\cite{arunkumar2017mcm,yin2018modular,schulte2015achieving, amdexascale,Hazelwood:HPCA2018} for accelerating the performance of general-purpose, large-scale parallel workloads~\cite{ryoo2008optimization,anderson2008general,krizhevsky2012imagenet}. At the same time, the GPU programming interface has matured, making GPUs accessible to amateur programmers. However, it is challenging to optimize and fine-tune the performance of general-purpose GPU programs without platform- and domain-specific knowledge. For example, programmers may be excessively cautious in their use of synchronization which incurs a performance penalty. \hlnew{It is well known that the lack of concurrency semantics adds to the challenge of optimizing parallel programs such as GPU codes~\cite{batty2015problem}. To continue the synchronization example, a compiler cannot perform additional optimization passes to eliminate unnecessary synchronization primitives
without concurrency semantics. Figure~\ref{fig:nw} illustrates the size of this effect for one of the Rodinia benchmarks: Eliminating unneeded synchronization nearly doubles the performance gain over O2 optimization for nw.}

Many research projects have investigated automated compilation optimization methods to reduce the burden on application programmers, including peephole~\cite{bansal2006automatic}, \hlnew{loop-unrolling using machine learning techniques~\cite{leather2009automatic}, auto-vectorization~\cite{mendis2019compiler}, auto phase ordering~\cite{haj2020autophase}}, link-time~\cite{van2005diablo}, and profile-guided optimization~\cite{pettis1990profile}, but additional efficiencies can be found by tailoring the binary to particular classes of inputs or particular architectures.
For example, STOKE is a stochastic program synthesizer that uses random search to explore the high-dimensional space of possible program transformations and is an example of this performance tuning approach~\cite{schkufza2013stochastic}. 
A related approach uses Evolutionary Computation (EC)
to morph existing code from the target program to code that improves run-time and other non-functional program properties. In this work we use EC, also known as Genetic Improvement, because earlier work showed that it scales well with program size~\cite{genprog-tse-journal,gp4energy}.
\hlnew{Some earlier work reports optimizations of specific GPU programs using EC~\cite{GP4cudaGzip,GP4cudaDNA, gp4cuda, GP4cudaRNA}, but these approaches are not easily extensible to large design spaces, in part because the representation of each program is customized and requires manual program translation before the search for optimizations can be performed.} 


We propose compiler post-pass performance tuning---{\bf GEVO}---\underline{G}PU code optimization using \underline{EVO}lutionary computation~\cite{GEVO.GI19, GEVO.WACI19}.
GEVO encodes optimization objectives, such as execution time, energy use or accuracy, in its fitness function and implements a set of mutation and recombination operators for GPU kernel transformations in the LLVM-IR representation.
We demonstrate GEVO first on the single-objective problem of reducing GPGPU kernel execution time (GEVO-default). Second, we show how GEVO can simultaneously tune GPU code to meet two independent objectives, such as performance and accuracy (GEVO-mO), using multi-objective search.

To assess the general applicability of GEVO, we evaluate on NVIDIA Tesla P100 using the Rodinia benchmark suite~\cite{pettis1990profile}, 
which covers a set of important application domains such as medical imaging and data mining. Each application provides unique characteristic in terms of parallelism, data sharing, memory access patterns, and so forth.  The results vary significantly across the Rodinia benchmarks, but on average
GEVO-default improves GPU kernel runtime performance by 49.48\% over the fully compiler-optimized baseline, and by as much as 412\% in one case. If output accuracy is relaxed to tolerate up to 1\% error, GEVO-mO can find kernel variants that outperform the baseline version by an average of 51.08\%. 

\hlnew{Although GEVO can be applied to any GPU program, we are particularly interested in machine learning programs, because machine learning tasks are computationally intensive and are by nature error-tolerant.  For example, deep neural networks often require hours or days to train a single model, and training time is often traded off against model accuracy~\cite{sgdlearning,recht2011hogwild,zhang2015deep,Hazelwood:HPCA2018,huang2017speed}. 
We present results for GEVO on supervised machine learning code from two production-level frameworks, ThunderSVM~\cite{thundersvm} and Caffe2~\cite{caffe2}, considering standard handwriting recognition, income prediction, and image classification datasets.}


To summarize, the key contributions of this paper are:
\begin {itemize}
    \item We present GEVO, a tool for automatically tuning the performance of GPU kernels represented in the LLVM intermediate representation (LLVM-IR) to meet multiple criteria. Our infrastructure scales to arbitrarily large program sizes.  We demonstrate GEVO on the single objective of runtime optimization and on the multi-objective optimization criteria of runtime and accuracy.
    
    \item Empirical evaluation on the Rodinia benchmark suite, which includes 13 applications covering a wide range of application domains, is performed. On average, GEVO improves kernel runtime by 49.48\% with the output fidelity enforced, or by 51.08\% if the output fidelity can be relaxed by 1\%. 
    
    \item Empirical evaluation of two machine learning kernels, using Thunder SVM and Caffe2 on standard machine learning benchmark datasets, is performed. In these experiments, model accuracy is interpreted as output fidelity.  Compared to the original baseline, we find that
    GEVO can improve kernel runtime performance by 1.79X to 3.24X. In most cases, these runtime improvements are achieved without loss of accuracy, and in some cases model accuracy actually improves. 

    \item In-depth analysis of GEVO optimizations identified several architectural-, domain-, and dataset-specific improvements. We provide explanations for many of the performance optimizations discovered by GEVO, such as eliminating conservative synchronization primitives (Section~\ref{sec:opt_if}), removing redundant {\tt store} instructions (Section~\ref{sec:opt_store}), reducing conditional executions (Section~\ref{sec:opt_if}), loop perforation (Section~\ref{sec:opt_loop}), memoization (Section~\ref{sec:opt_mem}), and algorithm improvements (Section~\ref{sec:opt_mo_rodinia} and~\ref{sec:ml_results}). 
    
    \item Multi-objective optimization: We demonstrate that when output fidelity is relaxed, solutions can be found that improve both optimization criteria---runtime and output fidelity---simultaneously. These optimization points are not accessible to the search when output fidelity is strictly enforced.

\end{itemize}

The remainder of the paper is organized as follows. Section~\ref{sec:related_work} provides relevant background and places the paper in the context of earlier work. Section~\ref{sec:Design} describes the GEVO design in detail; and Section~\ref{sec:experimental} describes the system environment and benchmarks we used to evaluate GEVO. Experimental results for GEVO-default and GEVO-mO are reported in Section~\ref{sec:result_default} and Section~\ref{sec:result_mo} respectively.  We discuss limitations and future directions in  Section~\ref{sec:discussion} and conclude the paper in Section~\ref{sec:conclusion}.

%% file: related.tex
\section{Related Work}
\vspace{-0.1cm}
\label{sec:related_work}

This section discusses related work from program synthesis, superoptimization, to evolutionary computation.  It also reviews the few earlier works that have applied EC to GPUs, and the growing body of work using EC to improve machine learning.
\subsection{Program Synthesis and Superoptimization}
\vspace{-0.1cm}
\label{sec:prog_synth}
\hlnew{Program synthesis methods automatically generate computer programs based on programmer-defined specifications. In some cases, the goal is to produce programs that run faster, which is known as superoptimization. Early work on superoptimization dates back to Massalin's superoptimizer~\cite{superoptimizer} in 1987, which exhaustively searched and tested a subset of the Motorola 68020 assembly instruction set against testing inputs, and synthesized the shortest instruction sequence, for a target function.}

\hlnew{Since the number of possible code sequences in most programming languages is enormous, finding an optimal sequence is usually intractable, and all recent program synthesis techniques use some form of search to cut down the search space. Such search algorithms can be roughly divided into two categories, deductive synthesis and inductive synthesis. Deductive synthesis encodes a given program into a Boolean formula and searches for a logically equivalent formula using an SAT/SMT solver. Developing such encodings is challenging and time-consuming, and significant research effort has been been devoted to making such processes more approachable to programmers, ranging from Rosette, a symbolic framework  language interpreter~\cite{10.1145/2509578.2509586, 10.1145/2594291.2594340} to Z3, an efficient SMT solver with C and Python binding~\cite{moura2008z3}. Synthesized programs using the deductive approach are provably correct and do not require verification. However, as the length of the computer program increases linearly, the size of the corresponding Boolean expressions grows exponentially. Therefore, program that are synthesized using the deductive approach are relatively small, and often have additional constraints. For example, the largest, loop-free program synthesized by Gulwani in 2011~\cite{gulwani2011synthesis} has only 16 lines in the pseudo assembly.}

\hlnew{Instead of represent the program's specification as a Boolean formula, inductive synthesis relies on the original program and a set of input/output examples (test cases). In this case, the search usually begins by sampling local deviations from the given program or by generating random code combinations from scratch.  Each variant is then checked against the test cases to verify functionality. Depending on the particular search method, new code recombinations are tried based on previous observations using statistical models such as Markov Chain Monte Carlo (MCMC) sampling, or heuristic search methods are used, as in the case of Evolutionary Computation (EC).
Schkufza et al. proposed STOKE~\cite{schkufza2013stochastic, schkufza2014stochastic, conditionalCorrectSO, soundLoopSO} using MCMC sampling for X86-64 assembly codes to improve run time. STOKE has the same overarching goal as GEVO, which is to search for program optimizations without guaranteeing exact program semantics as deductive synthesis requires. However, STOKE does not naturally scale up to large code sizes because it considers the entire X86-64 instruction set, even vector instructions, and it
focuses on synthesizing instruction sequences from scratch. EC differs from STOKE in two ways: It modifies existing programs using existing instructions, and it can scale to arbitrarily large program sizes (over a hundred thousand instructions~\cite{gp4energy}), whereas the test programs in STOKE are only a few hundred instructions~\cite{soundLoopSO}. }

\hlnew{We elected to use EC primarily because of its scalability to realistic program sizes. In addition to scalability,   an inductive method was appropriate because many applications on parallel accelerators, e.g., image processing and machine learning, do not require an exact result. Instead, these applications often have domain specific metrics for assessing an acceptable result. This feature of many GPU program mitigates the requirement to preserve exact program semantics.} 

\subsection{Evolutionary Computation (EC)}
\vspace{-0.1cm}

Earlier work developed EC methods to improve computer programs, e.g., to
automatically repair bugs in legacy software~\cite{study4AutoProgRepair, fixes-suggest,genprog, Weimer:gp4patches,GP4AutoSWrepair}, and this class of applications is sometimes referred to as \textit{genetic improvement}. Transitions to industrial practice include
Facebook's SapFix tool~\cite{sapfix} and the Janus Manager
deployment~\cite{haraldsson17}.  
Although most work has been conducted at the source-code level using abstract syntax trees, similar methods 
have been applied to assembly programs~\cite{schulte2013} and object
code~\cite{SchulteEtAl2015a}. 

Subsequent analysis showed that
the applied mutations often have no observable effect on program
behavior~\cite{SWmutationRobust,Baudry2015,bruce15energy,haraldsson17a,Veerapen17}.
These \emph{neutral} mutations occur frequently ($20 - 40\% $ of the time), even 
when the mutations are restricted to sections of code covered by the tests.  
Although it is surprising that the rate of neutral mutations is so high, equivalent mutations are well-known in mutation testing, e.g.,~\cite{MadeyskiEquivMutants}.  
These results suggested the possibility of using EC to optimize non-functional properties of software by finding modifications that are
neutral with respect to the test suite but improve the non-functional property in question.

White et al. proposed the idea of using EC to improve program
performance~\cite{GP4ProgImprovement}, and Schulte et al. achieved
significant energy reductions for several Parsec
benchmarks~\cite{gp4energy}.  
Bruce et al. applied a similar technique for MiniSAT to reduce energy consumption~\cite{bruce15energy}. Other works~\cite{burles2015object, manotas2014seeds} constrain EC's search space for improved energy consumption of Java programs by asking users to provide predefined locations or equivalent functions or class implementations.
These, and several subsequent works~\cite{bruce2018approximate, dorn17energy}, demonstrate the
potential for stochastic search methods such as EC to improve a program's performance or
energy efficiency through machine- or architecture-specific
optimizations.  However, these methods are not yet mature or carefully analyzed.  In contrast with our work, they focus on energy reduction rather than run-time, typically target the CPU; and their general applicability is not well understood.  The results reported here address these limitations.

\subsection{Evolutionary Computation for GPU Programs}
\begin{wrapfigure}{R}{0.5\textwidth}
	\center
	\includegraphics[width=0.48\textwidth]{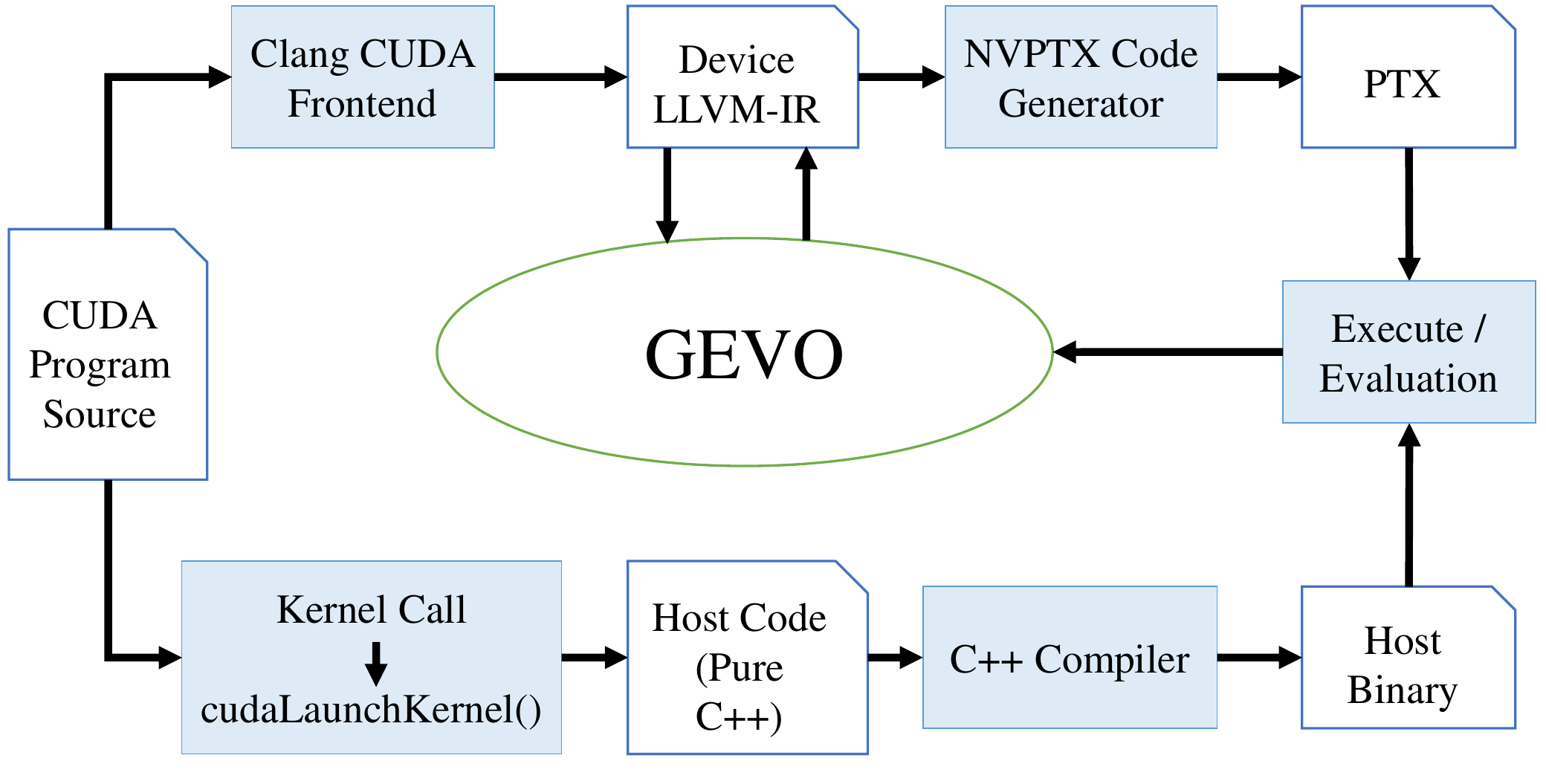}
	\vspace{-0.5cm}
	\caption{GEVO in the LLVM/Clang compilation flow.}
	\label{fig:cuda-compileflow} 
\end{wrapfigure}

There is some previous work applying EC to GPU kernels, including a graphics
shader program~\cite{gp4shader}. This work began with a basic lighting
algorithm and used EC to gradually modify the shader program into a
form that resembles an advanced algorithm proposed by domain specific
experts.  In the GPGPU domain, Langdon et al. used EC to reduce
the runtime of a CUDA program, reporting results for two specific programs:
\texttt{gzip}~\cite{GP4cudaGzip} and an RNA analysis
program~\cite{GP4cudaDNA}.  This prior work targets a
single program operating in a specific domain, and the
methodology used in~\cite{GP4cudaGzip,GP4cudaDNA, gp4cuda, GP4cudaRNA}
represents the program object as a custom-designed,
line-based Backus Normal Form (BNF) grammar.  We seek a method that is
generalizable across multiple programs with minimal manual
intervention and uses modern tooling, which is Clang/LLVM.

Clang/LLVM is a widely-used, multi-language, and highly modular
compiler infrastructure. Schulte's thesis is the only work we are aware of that has experimented with evolving the LLVM
intermediate representation (LLVM-IR)~\cite{schulte2014dissertation}, but this was a preliminary proof-of-concept rather than a robust implementation, and no significant experiments were conducted.  Now that Clang/LLVM supports CUDA
compilation, it is feasible to compile GPU kernels into LLVM-IR, but
this has only been available since 2016~\cite{gpucc}. Thus, we adopt the Clang/LLVM infrastructure
for GEVO including the LLVM-IR, as shown in Figure
\ref{fig:cuda-compileflow}. This avoids developing novel parsing and
syntax manipulating infrastructure, but it introduces new challenges
for implementing the basic mutation and recombination operations.

\subsection{Genetic Improvement of Machine Learning (ML)}
\label{sec:related_work:ML}
\vspace{-0.1cm}
While EC can be applied to any computer program, machine learning is a particularly attractive domain, because of its popularity and its high computational cost.
Moreover, most ML applications can be accelerated using GPUs. Although no prior work is known that uses EC to accelerate ML programs on GPUs, EC has often been used to improve neural network designs and to optimize weights.  This work dates back at least to an 1989~\cite{montana1989training} paper that used EC
train a neural network. 
The most established and popular approach in this domain is NEAT~\cite{neat}, proposed by Stanley et al. in 2002, which uses EC to simultaneously learn the neural network connection topology and the weight of each neuron. 
The original NEAT design performed well in a comparatively small yet homogeneous neural network where the network consists of common neurons, and the following works expanded its scalability to larger networks and more complex tasks~\cite{stanley2009hypercube, verbancsics2011constraining, real2017large}. 

More recently, convolution neural networks (CNNs) have achieved extraordinary performance in image classification tasks by providing additional convolution layers as filters. 
These layers are used to determine important spatial pattern in the image so that the number of features can be reduced before being fed into a traditional neural network. Many approaches for identifying good architectures (topologies) for CNNs have been proposed~\cite{pham2018efficient, bender2018understanding, liu2018progressive, kandasamy2018neural, liu2017learning, liu2017hierarchical, xie2017genetic, zoph2018learning}, and these have outperformed manually designed architectures in several tasks. Similar to NEAT, Reak et al. proposed using EC to design CNNs in a limited search space of convolution layers composed by common arithmetic operations~\cite{real2019regularized}. This work achieves state-of-the-art classification performance on the ImageNet dataset compared to other network architecture searches, which use random search and reinforcement learning. 

\hlnew{
In state-of-the-art machine learning programming frameworks, deep learning models are represented as computational graphs of various types of operators. 
This exposes opportunities for operator-level optimization, such as operator fusion, using domain-specific compilers. For example, XLA~\cite{xla} is developed for TensorFlow~\cite{tensorflow}, Glow~\cite{rotem2018glow} for PyTorch~\cite{paszke2017automatic}, TVM~\cite{chen2018tvm} for MXnet~\cite{chen2015mxnet}, and so forth.  Domain-specific compilers can perform further optimization when lowering high-level, neural-network operators onto machine-specific implementations using optimized libraries. These optimizations differ from the aforementioned neural architecture searches in that the functional behavior of a given network is preserved. Built on top of the prior superoptimization approach~\cite{superoptimizer}, TASO~\cite{taso} was recently proposed to automatically optimize the computational graph. Given a small computational graph, TASO enumerates all combinations of operator implementations and selects the operator graph implementation that minimizes runtime. The functional behavior of the original graph is preserved with satisfiability verification using a SAT solver. 
TASO shares the scalability limitation discussed earlier for deductive program synthesis (Section~\ref{sec:prog_synth}), and the referenced work does not scale beyond graphs of size 4 operators. 
}

\hlnew{
Our approach to optimizing NNs is complementary to this earlier work. GEVO explores joint optimization opportunities by 
(1) discovering better-performing operator implementations and (2) changing neural network architectures, which extends previous work (Section~\ref{sec:caffe2_results}).
}


%% file: design.tex
\section{GEVO Design}
\vspace{-0.1cm}

\label{sec:Design}

\def\BState{\State\hskip-\ALG@thistlm}
\makeatother
\begin{wrapfigure}{R}{0.53\textwidth}
\begin{minipage}{\linewidth}
\begin{algorithm}[H]
\small
\caption{The main loop of GEVO. }\label{pseudocode}
\hspace*{\algorithmicindent} \textbf{Parameter:} \textit{PopSize, CrossRate, MutateRate, InitDist} \\
\hspace*{\algorithmicindent} \textbf{Input: } \textit{GPU kernel Program, P} 
\begin{algorithmic}[1]
\State $\textit{pop} \gets \text{Initialize(\textit{PopSize, P})}$
\ForAll{\textit{individual} in \textit{pop}}
    \State $\text{Mutate(\textit{individual})*\textit{InitDist}}$
\EndFor
\State $\textit{rank} \gets \text{NonDominatedSort(\textit{pop})}$
\While{not interrupt}
    \State $\textit{offspring} \gets \text{SelTournament(\textit{pop, rank, PopSize})}$
    \State $\textit{elites} \gets \text{SelBest(\textit{pop, rank, PopSize} /4)}$\\
    
    \For{every 2 individual (\textit{ind1, ind2}) in \textit{offspring}}
        \If{random() < \textit{CrossRate}}
            \State $\text{Crossover(\textit{ind1, ind2})}$
        \EndIf
    \EndFor
    \ForAll{\textit{individual} in \textit{offspring}}
        \If{random() < \textit{MutateRate}}
            \State $\text{Mutate(\textit{individual})}$
        \EndIf
    \EndFor \\
    
    \State $\textit{rank} \gets \text{NonDominatedSort(\textit{elites + offspring})}$
    \State $\textit{pop} \gets \text{SelBest(\textit{\textit{elites + offspring}, rank, PopSize})}$
\EndWhile
\end{algorithmic}
\end{algorithm}
\end{minipage}
\end{wrapfigure}

We propose {\it GEVO}---a tool for automatically improving kernel implementations for GPUs. GEVO enables \underline{G}PU code optimization using \underline{EVO}lutionary computation. It is a post-pass performance tuning approach to optimizing GPGPU kernel implementations.

GEVO takes as input a GPGPU program, user-defined test cases that specify required functionality, and a fitness function to be optimized. GEVO attempts to maximize the fitness function by evolving and evaluating mutated kernel variants in an iterative population-based search. Figure~\ref{fig:cuda-compileflow} presents GEVO in the context of the LLVM/Clang compilation flow. 
Kernels in a GPGPU program that will run on GPU are first separated and compiled into the LLVM intermediate representation (LLVM-IR) using the Clang compiler. GEVO then takes kernels in LLVM-IR format as inputs, modifies them to produce different kernel variants, and translates the variants into PTX files. The host code running on CPU is modified to load the generated PTX file into the GPU.  GEVO then evaluates how the kernel variant performs as defined by the objectives encoded in the fitness function.


As shown in Algorithm~\ref{pseudocode}, the search begins with an initial population of PopSize individuals (LLVM-IR kernel variants) that is formed by taking the original program, making PopSize copies and applying random mutations to each (Line 3 where InitDist, the number of mutations applied to each individual, defaults to 3), giving the initial population some diversity.  GEVO then forms the next generation of individuals by ranking individuals according to the objectives, recombining instructions between kernel variants ({\it Crossover}), and randomly adding, deleting or moving instructions in each variant ({\it Mutation}). Finally, GEVO compares the new variants to a set of elites retained from the previous generation ({\it Selection}), eliminating low-fitness individuals and retaining those with higher fitness for the next generation.  The next few subsections give details of how we implemented these operations for GPGPU optimization. 

\subsection{Individual Representation}
\vspace{-0.1cm}

GI methods typically use either a program-based (each variant consists of the entire program) or a patch-based (each variant is a list of mutations (edits) applied to the original program) representation.
For large programs, the patch-based representation is convenient because it is more space efficient.  GEVO uses both representations.  That is, each individual kernel variant contains both the LLVM-IR code and the set of the mutations that produced it from the original. 

This design decision relates to the many data dependencies built into the LLVM-IR.  Because of the repair process that is required after most mutations, it would be expensive to reapply all the mutations for a variant each time it is evaluated.  Similarly, crossover exchanges subsections of the kernel code.  Doing this naively can break a large number of data dependencies, so it is more efficient to implement crossover using the patch representation.  Because the number of mutations applied to any kernel variant tends to be low, and because kernels are relatively small-sized programs, the memory requirement of storing both representations is reasonable.

\subsection{Fitness Evaluation} 
\label{sec:design:evaluation}
\vspace{-0.1cm}

Although GEVO can optimize any desired fitness function, we first focus on the problem of reducing kernel execution time ({\it GEVO-default}). In this case, the fitness function is simply the runtime of the kernel variant \hlnew{without tolerating any output variation}. When we consider approximate computing, where output accuracy can be relaxed to improved execution time, we include output accuracy in the fitness function as multi-objective ({\it GEVO-mO}), i.e., $argmin(time, error)$. 

Using these fitness criteria each kernel variant is evaluated by running it against all available test cases. To protect against overfitting, we also evaluate at the end of the search against a set of held-out test cases, generated randomly. The fitness value is reported as a vector corresponding to the number of objectives. Each element in the vector is a single scalar value, i.e., the mean performance on that objective across the test cases (see Section~\ref{sec:error_metric}). 

\subsection{Selection}
\vspace{-0.1cm}

\begin{figure}[t]
\begin{minipage}[h]{0.47\textwidth}
	\includegraphics[width=\textwidth]{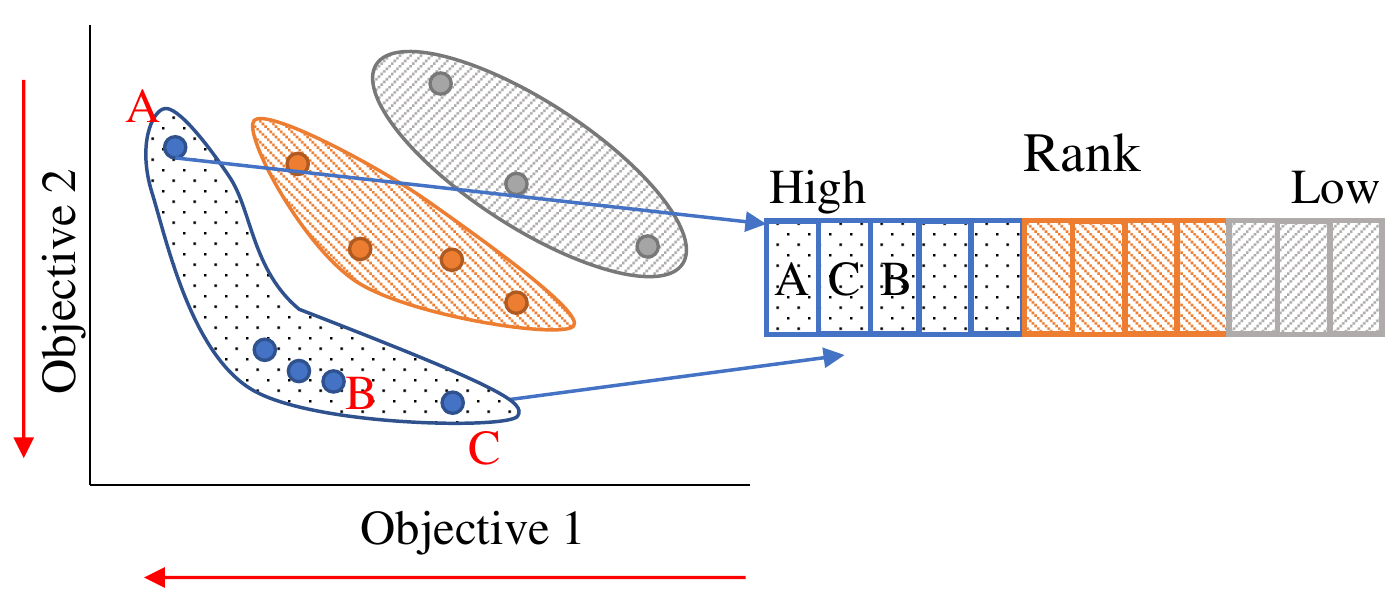}
	\vspace{-0.5cm}
	\caption{Non-dominated sorting with a crowding method to enforce diversity.	}
	\label{fig:nsga2} 
\end{minipage}\hfill
\begin{minipage}[h]{0.47\textwidth}
	\includegraphics[width=\textwidth]{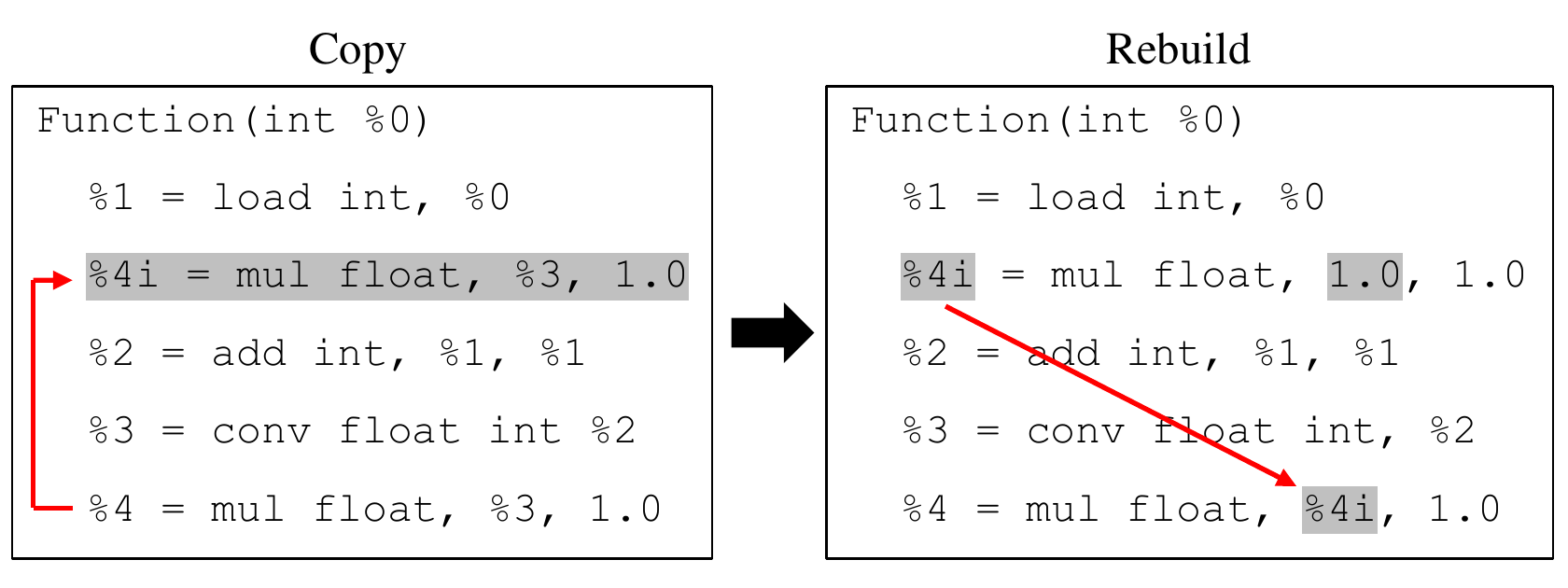}
	\caption{Mutate-Copy: Operand dependency is rebuilt to preserve LLVM-IR program validity. Since LLVM-IR is strongly typed, the constant value 1.0 is used if no other value in the requested type is available. }
	\label{fig:mutation_rebuild} 
\end{minipage}\hfill
\end{figure}

GEVO uses the Non-dominated Sorting Genetic Algorithm (NSGA-II)~\cite{nsga-ii} to select individuals according to multi-objective fitness criteria. Figure~\ref{fig:nsga2} illustrates a set of kernel variants, plotted according to two dimensions of the fitness function, say error and run-time, where the goal is to minimize both objectives, retaining individuals that represent the best tradeoffs between the two objectives (shown in blue in the Figure). 
NSGA-II uses Pareto dominance, where individual $i$ is said to dominate individual $j$ if $i$ is better than $j$ on at least one objective and no worse on the others.

NSGA-II calculates the Pareto fronts, and ranks individuals according to which front they belong.  Then a crowding factor is calculated for each individual based on the density of other individuals along its Pareto front, and these two values are combined to produce a single fitness value for each kernel variant.  See~\cite{nsga-ii} for details. Finally, based on this single fitness value, NSGA-II uses a popular EC selection method known as \textit{tournament selection}~\cite{miller1995tournament} to choose kernel variants for the next generation (Line 6 of Algorithm~\ref{pseudocode}).
GEVO retains the top quarter of the population at time $t$ and copies it unchanged to the population at time $t + 1$ (Lines 7, 16, and 17 of Algorithm~\ref{pseudocode}), a method known as \textit{elitism}~\cite{baluja1995elitism}.  It then chooses the remaining $3/4$ of the population using the tournament selection.

\subsection{Mutation Operators}
\vspace{-0.1cm}

Mutation modifies the linear array of instructions stored for each variant using one of the following operations: 
\begin{itemize}
\item {\bf Mutate-Copy}: Duplicate an instruction and insert it in another location.
\item {\bf Mutate-Delete}: Remove an instruction.
\item {\bf Mutate-Move}: Move an instruction to a different location.
\item {\bf Mutate-Replace}: Replace an instruction with another instruction. This can occur either at the instruction or the operand level. In the second case, a single operand is replaced with another operand.
\item {\bf Mutate-Swap}: Swap the location of two instructions.
\end{itemize}

Since the LLVM-IR is based on Static Single Assignment (SSA) where each variable (like \texttt{\%0, \%1} in Figure~\ref{fig:mutation_rebuild}) can be assigned only once at creation, our mutations are likely to create invalid programs by breaking the SSA constraint. Thus, we introduce an extra repair step. As illustrated in Figure~\ref{fig:mutation_rebuild}, the instruction \texttt{mul} is copied, and we see that the first operand relies on \texttt{\%3} which is invalid in the new location. GEVO repairs it with the constant 1.0 as the two existing values (\texttt{\%0, \%1}) are not of the proper type.  To our knowledge, only one other work~\cite{schulte2014dissertation} 
has attempted to design mutation operations for SSA.  GEVO implements similar mutations to this earlier work, although our mutations repair SSA dependencies more robustly, and we have introduced two new mutation operations.

The operators Mutate-Copy and Mutate-Move insert new instructions, which has no effect unless a subsequent instruction can use the result of the inserted instruction. Figure~\ref{fig:mutation_rebuild} illustrates how GEVO enforces this by changing the first operand of the fifth instruction to use the value from second instruction. This instruction was selected because its types agree with the mutated instruction. \hlnew{
In addition to the type checking shown in the above repair procedure, dominator analysis weeds out values if the creator and the user of that value are in separated basic blocks that do not share the same execution path.
}

As depicted in Line 14 of Algorithm~\ref{pseudocode}, when mutation is called, one of the aforementioned mutation operations is selected randomly (with equal probability) and applied as an edit to generate a new kernel variant. Since GEVO does not use domain-specific knowledge to select a mutation or rely on program semantics, we immediately evaluate the individual \hlnew{by running the available test cases, as Section~\ref{sec:design:evaluation} describes,} and eliminate invalid modifications (sanity check). 
Mutation is iteratively applied to the same individual until a valid kernel variant is identified. Depending on the kernels, the acceptance rate of any single mutation is typically 5\% - 30\%. 

\subsection{Crossover}
\vspace{-0.1cm}
\label{sec:crossover}

\begin{wrapfigure}{R}{0.55\textwidth}
    \centering
	\includegraphics[width=0.53\textwidth]{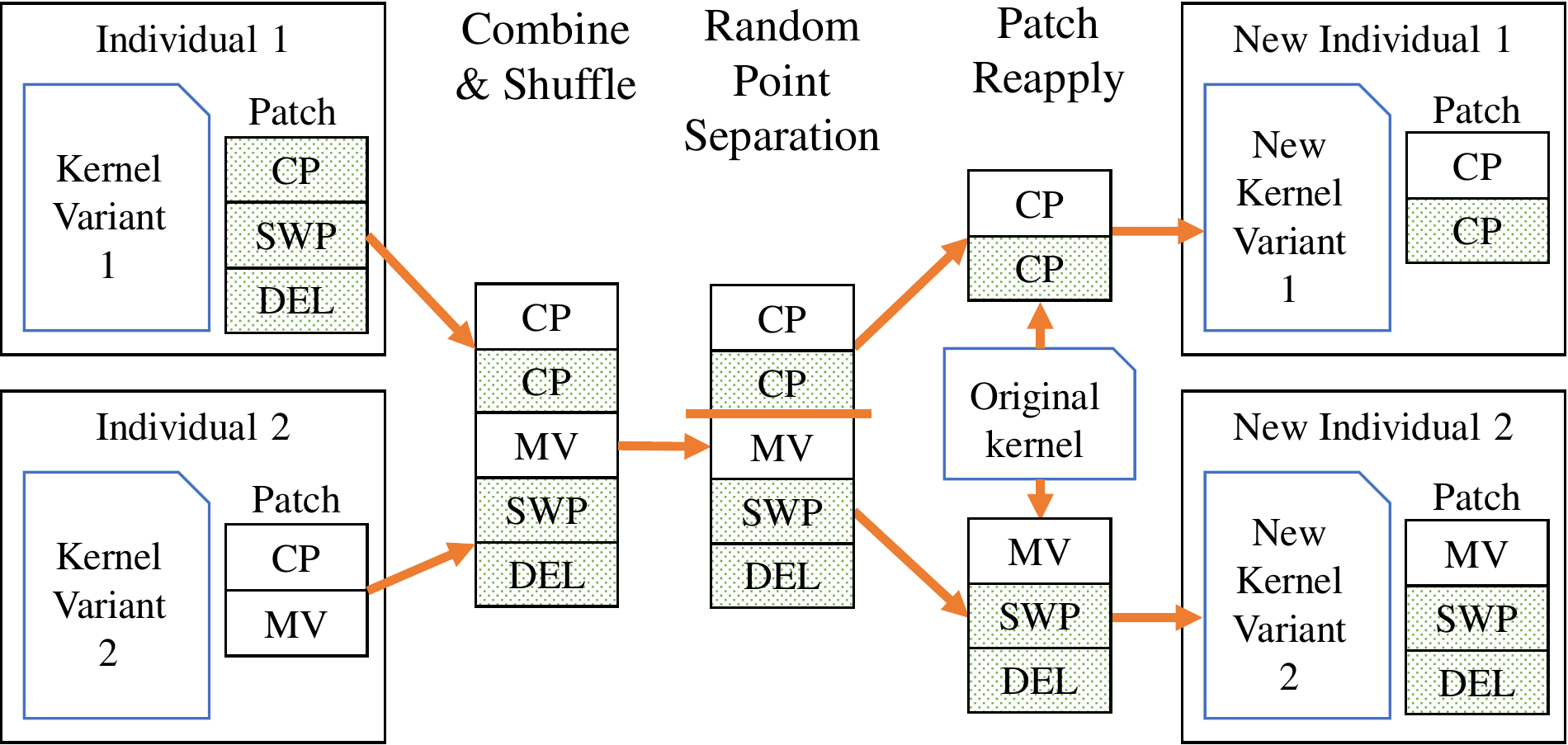}
	\vspace{-0.4cm}
	\caption{One-point messy crossover. }
	\label{fig:crossover} 
\end{wrapfigure}

GEVO uses the patch-based representation for crossover, because combining two random program slices would require more extensive repair. 
GEVO implements one-point messy crossover, which combines shuffle~\cite{shufflecrossover} and variable-length~\cite{Lee00variablelength} crossover operations. Figure~\ref{fig:crossover} illustrates the process. Beginning with a list of mutations (edits) associated with each individual, GEVO combines them into a single sequence, shuffles the sequence, and randomly separates it back into two distinct patch sequences. Finally, GEVO reapplies each patch sequence to the original GPU kernel and generates two new individuals. This form of crossover was selected because it generates a wide diversity of recombinations from a minimal number of mutations, and our mutations are relatively expensive.

Similar to mutation, after crossover, each new individual is evaluated to test if the combinations are valid. Otherwise, GEVO repeats the process until it finds a successful recombination. The acceptance rate of crossover is as high as 80\% because each individual mutation has already been validated.

%% file: experimental.tex
\section{Experimental Setup}
\vspace{-0.1cm}
\label{sec:experimental}

This section describes our experimental setup for the
empirical evaluation on real GPUs.

\subsection{Infrastructure and Configurations}
\vspace{-0.1cm}

We developed GEVO using an existing python framework for evolutionary algorithms, called DEAP~\cite{deap}, implementing the genetic operators described in Section~\ref{sec:Design}, and integrating them into the DEAP framework.\footnote{The GEVO code is available at \url{https://github.com/lioujheyu/cuda_evolve}} We instrument the LLVM compiler (LLVM 8.0) to implement the mutation operations in C++ as a LLVM pass. For each optimization, GEVO was given a 48-hour wall-clock budget. We evaluate GEVO using NVIDIA Tesla P100 GPUs, under CUDA 10.1 and NVIDIA GPU driver 418.87. The Nvidia profiler (nvprof) was used to collect kernel execution time, which became the runtime metric used by the fitness function. In our experiments, nvprof introduced no overhead to kernel execution time but only to application execution time which is not the optimization target in this study. Also, the collected kernel execution time consistently varies less than 1\% through multiple profiling run on the same GPU kernel.

All GEVO experiments were conducted with population size of 256, crossover rate of 80\% (80\% of individuals in population are selected for crossover), and a mutation rate of 30\% (every individual has 30\% chance to receiving one mutation). The 48-hour budget for each GEVO run translates into a variable number of generations (shown in the last column of table~\ref{table:benchmarks}), depending on the program, the test cases, and the success rate of the mutation operation. For our experiments, the number of generations ranged from a low of 12 to over 80.  For example, for the NN benchmark, GEVO spent the majority of its time searching for valid mutations and was able to complete only 18 generations within 48 hours. 
We speculate that in cases where GEVO was unable to find useful optimizations it is partially because the runs did not complete enough iterations of the search, and providing a larger search budget could improve results for these programs.

\begin{table}
\small
\centering
\caption{Benchmarks used for evaluation.}
\vspace{-0.2cm}
\label{table:benchmarks}

\begin{tabular}{|l l r r|} 
 \hline
 & & GPU Kernel & \\
 Application & Abbr. & Line of LLVM-IR & Generation\\ 
  \hline\hline
 Breadth first Search & bfs & 72 & 18\\ 
 B+Tree & b+t & 168 & 63\\
 CFD Euler3D & cfd & 1079 & 53 \\
 Gaussian elimination & gau & 186 & 29\\
 Heart Wall & hw & 3806 & 36\\
 Hotspot & hot & 189 & 28\\
 LU decomposition & lud & 2491 & 81\\
 Nearest Neighbor & nn & 32 & 18\\
 Needleman-Wunsch & nw & 715 & 21\\
 Particlefilter & pf & 1442 & 55\\
 Pathfinder & path & 109 & 25\\
 SRAD\_v2 & sv2 & 446 & 16\\ 
 Stream Cluster & sc & 231 & 12\\
  \hline
  Handwriting recognition (C=5, g=0.05) & mnist & (c\_smo\_solve) 256 & 27\\
  Income prediction (C=32, g=0.0078125) & a9a & (c\_smo\_solve) 256 & 61\\
  \hline
  Image classfication & cifar-10 & (momentumSGD) 39 & 15\\
  \hline
\end{tabular}
\end{table}

\subsection{Benchmarks}
\label{sec:benchmarks}
\vspace{-0.1cm}
Table~\ref{table:benchmarks} summarizes the benchmarks we used to evaluate GEVO: the Rodinia benchmark suite and ML workloads on ThunderSVM and Caffe2.
Rodinia includes a wide range of general-purpose deterministic workloads for heterogeneous computing, representing diverse parallel communication patterns, synchronization techniques and power consumption. ML models are a natural application for exploring accuracy/efficiency tradeoffs because the algorithms are
intrinsically error-tolerant, and they require significant time overhead for training, e.g., the training time of state-of-the-art language translation models is on the order of days~\cite{Hazelwood:HPCA2018}. 

For the first set of benchmarks, we validate optimized kernel variants using the default inputs provided with the Rodinia benchmarks. For each benchmark, we then generate additional tests by randomly generating three sets of input values using the Rodinia built-in input generator. Depending on the benchmark, each input set contains from tens of thousands to millions of input values. Each test is run with the original, unmodified kernel and its output is used as an oracle to validate the output of the candidate kernel variants. GEVO rejects variants that fail to produce outputs that are identical to the oracle (GEVO-default) or exceed the default 1\% error tolerance (GEVO-mO). 
After evolution, we validate the highest fitness kernel variant found during the search on held-out tests.  We generate the held-out tests by rerunning the test-generation process.  This step helps ensure that GEVO does not overfit the kernel to the existing test cases during the evolution.

For the ML benchmarks, we focused on a Support Vector Machine (SVM) and Stochastic Gradient Descent.
Because GEVO searches the optimization space at the granularity of instructions, it requires full access to the intermediate representation and the corresponding optimized library implementations. 
This consideration led us to a supervised ML framework, ThunderSVM, which is a support vector machine library that is fully open-sourced and optimized for GPU implementation.

We evaluate GEVO on ThunderSVM (refered to as SVM) with two standard datasets: handwriting recognition using MNIST~\cite{mnist} and income prediction using a9a~\cite{a9a}. We downloaded the datasets from libsvm~\cite{libsvm}'s data repository, which consists of 60,000 training and 10,000 testing data samples for MNIST, and 32,561 training and 16,281 testing samples for a9a. Additionally, we used the MNIST large dataset, which contains 8,000,000 image samples, solely for the post-evolution evaluation. Specifically, we asked GEVO to optimize the {\tt c\_smo\_solve} kernel for both training time and inference prediction accuracy of the trained model. We present the results in Section~\ref{sec:svm_results}. 

Many popular deep learning frameworks, like TensorFlow~\cite{tensorflow}, PyTorch~\cite{paszke2017automatic}, and Caffe2~\cite{caffe2}, rely heavily on the NVIDIA closed source cuDNN library to drive the GPU and cannot be directly targeted by GEVO.  However, a few frameworks maintain a small custom CUDA implementation when required functionality has no direct mapping from the cuDNN library. For example, in Caffe2, stochastic gradient decent with momentum (momentumSGD) is custom implemented as a CUDA kernel and open sourced within Caffe2 source repository. This provides an opportunity for us to include Caffe2 in our evaluation.

We evaluate Caffe2 using an 18-layer residual neural network (refered as ResNet18) to perform image classification against the CIFAR-10 dataset~\cite{cifar10} with 50000 training and 10000 testing images. We used GEVO to optimize the momentumSGD kernel because it is the only major operator used in ResNet that is open-sourced to us. This kernel updates the weight of the neural network based on the loss function, evaluating the difference between the true label and predicted label. Since the search space is constrained to a single operator, we include this application to demonstrate GEVO's capability and do not attempt a sophisticated solution for image classification. The results are given in Section~\ref{sec:caffe2_results}.  

\subsection{Error Metric}
\vspace{-0.1cm}
\label{sec:error_metric}

For the Rodinia benchmarks, error represents the maximum difference between outputs produced by the unmodified, original kernel implementation and that of GEVO-mO, across all tested inputs. 
Kernel variants are eliminated if the error rate of any test case exceeds the prespecified threshold, i.e. 1\%. 

For both SVM and ResNet18, we consider the runtime to train the model and the accuracy of the trained model's performance. For SVM, we use two-fold cross validation on the training dataset to report the error to GEVO during optimization. In ResNet18, we also report the training error for GEVO when the  model is trained for three epochs, which  shortens the training time to one minute. However, even with this simplification, GEVO required seven days to search for 20 generations, which is a low number of generations for an evolutionary search. The testing dataset regardless of which framework and application is only for reporting the testing error and is never be presented to GEVO.

Similar to Rodinia, the ML kernel variants are rejected if the training error exceeds the error achieved from the original kernel by the 1\% threshold. For example, if the the unmodified kernel achieves 3\% training error, then a GEVO kernel variant with 4.1\% training error will be rejected, and another one with 3.9\% training error will be accepted.
In ResNet18, we set the threshold to 10\% because we trained for only three epochs, which is generally not long enough for the model to converge.  Thus, the training error is noisier with this training regime, so we require a more generous error tolerance.

%% file: result.tex
\section{Evaluation of GEVO-default}
\label{sec:result_default}

\definecolor{backcolour}{rgb}{0.95,0.95,0.92}
\definecolor{codegreen}{rgb}{0,0.6,0}
\definecolor{codegray}{rgb}{0.5,0.5,0.5}
\definecolor{codepurple}{rgb}{0.58,0,0.82}
\definecolor{backcolour}{rgb}{0.95,0.95,0.92}

\lstdefinestyle{mystyle}{
    language=C++,
    morekeywords = {__shared__, __syncthreads},
    otherkeywords = {__shared__, __syncthreads},
    backgroundcolor=\color{backcolour},
    keywordstyle=\bfseries\color{blue},
    numberstyle=\ttfamily\tiny\color{codegray},
    commentstyle=\color{codegreen},
    basicstyle=\tiny\ttfamily,
    numbers=left,
    frame=single,framexleftmargin=3em,
    stepnumber=1,
    showstringspaces=false,
    tabsize=1,
    breaklines=true,
    breakatwhitespace=false,
    xleftmargin=.20\textwidth, xrightmargin=.20\textwidth
}

\lstdefinestyle{2column}{
    language=C++,
    morekeywords = {__shared__, __syncthreads},
    otherkeywords = {__shared__, __syncthreads},
    backgroundcolor=\color{backcolour},
    keywordstyle=\bfseries\color{blue},
    numberstyle=\ttfamily\tiny\color{codegray},
    commentstyle=\color{codegreen},
    basicstyle=\footnotesize\ttfamily,
    numbers=left,
    frame=single,framexleftmargin=2em,
    stepnumber=1,
    showstringspaces=false,
    tabsize=1,
    breaklines=true,
    breakatwhitespace=false,
    xleftmargin=2.5em, xrightmargin=.0\textwidth
}

\begin{figure}[t]
	\center
	\includegraphics[width=0.6\linewidth]{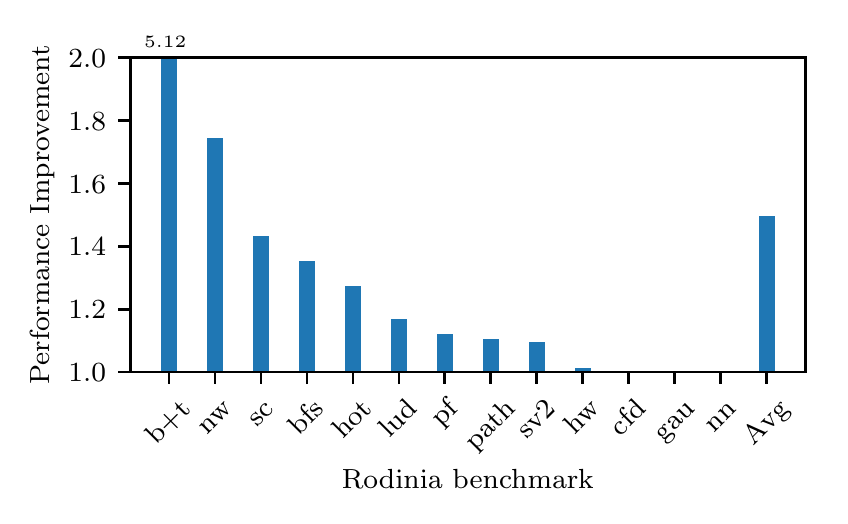}
	\caption{Normalized performance improvement over the default baseline with full compilation optimization for GEVO-default in the Rodinia Benchmark. 
	(For example, the 1.27X improvement in hot reduces runtime from 1.07 seconds to 1.07/1.27 = 0.84 seconds.)
	}
	\vspace{-0.3cm}
	\label{fig:result_rodinia_default} 
\end{figure}

We first present the results of our experimental evaluation on the Rodinia benchmark suite, with the goal of evaluating GEVO's applicability across a variety of programs (GEVO-default). Next, we take an in-depth look at the most common \textit{architecture-specific} and \textit{application-specific} optimizations that GEVO discovered (Section~\ref{sec:arch-app-optimizations}).  In Section~\ref{sec:ml_results}, we consider \textit{dataset-specific} optimizations.

Figure~\ref{fig:result_rodinia_default} reports the overall performance improvement for GEVO-default by comparing to the default baseline with full compilation optimization for the Rodinia benchmarks. GEVO-default improves the performance of the Rodinia benchmark suite by an average of 49.48\% and by as much as 412\% for {\tt b+t}. As figure~\ref{fig:result_rodinia_default} shows, there is significant variability in the achieved improvement among programs, including three, \texttt{cfd}, \texttt{gau} and \texttt{nn}, for which we found no optimizations. There are several possible explanations for this variability. It could be a feature of the program itself, it could be that we did not let GEVO search long enough, or perhaps the program was perfectly optimized by the original programmer.  However, this variability is consistent with results reported using evolutionary methods on the related problem of reducing energy use by assembly programs~\cite{gp4energy}.

\subsection{Architecture-specific Optimizations}
\vspace{-0.1cm}
\label{sec:arch-app-optimizations}

GEVO discovered several different optimizations related to GPU architecture design in the evolved Rodinia Benchmarks, including 
synchronization issues and memory order issues. Some of these optimizations arise from a combined effect of the architecture and the particular application/algorithm. 

\subsubsection{Eliminating synchronization primitives}
\label{sec:opt_sync}

\begin{wrapfigure}{R}{0.50\textwidth}
\lstset{style=2column}
\begin{lstlisting}
__shared__  int temp[...][...];
__shared__  int ref[...];
int tid = threadId.x;

ref[tid] = referrence[...];
__syncthreads();
temp[tid+1][0] = matrix_cuda[...];
__syncthreads();
temp[0][tid+1] = matrix_cuda[...];
__syncthreads();

for (int i=0; i<BLOCK_SIZE; i++)
  temp[tid][tid] = 
    temp[i][0] + temp[0][i] + ref[i];
\end{lstlisting}
\vspace{-0.4cm}
\caption{Simplified code snippet from \texttt{nw} with conservative {\tt syncthread()} calls.}
\label{fig:not-needed-sync}
\end{wrapfigure}

One of the most common GEVO optimizations removed synchronization primitives, specifically \texttt{syncthread()} calls in CUDA. For example, when a programmer wants to exchange data between threads in a thread block through the shared or global memory, synchronization primitives are used to synchronize the progress of GPU threads.
There are multiple reasons why a \texttt{syncthread()} call might not be required and could be removed without damaging the application. We give several examples, taken from GEVO runs on the Rodinia benchmarks.

\underline{\tt \bf nw}: Figure~\ref{fig:not-needed-sync} shows a simplified code snippet taken from \texttt{nw}. The \texttt{syncthread} function is used three times in this particular kernel (Lines 6, 8, and 10). The first two \texttt{syncthread} calls (Lines 6 and 8) can be eliminated because neither \texttt{ref} nor \texttt{temp} are read before a new value is written into the same memory location.  It appears that the programmer was
unnecessarily conservative in this case, which increased performance cost without additional semantic value.
Most of the performance improvements discovered by GEVO for {\tt nw} eliminated similarly overly-conservative uses of \texttt{syncthread}. Such optimizations are, of course, somewhat risky in general.  However, they illustrate the value of tailoring a kernel for exactly the workloads it will experience.  As part of a wide deployment, additional post-hoc methods, such as test-case generation or program analysis, could be employed to double-check that specific optimizations are indeed safe under the relevant use cases.

\underline{\tt \bf lud}: Other synchronization-related optimizations found by GEVO are architecture-specific and concern scope. In a GPGPU application, a massive number of parallel threads are created to execute the same piece of code in a kernel. At runtime, multiple threads form a {\it thread-block} or a {\it cooperative thread array}. A thread-block is the basic unit of
execution---all threads within a thread-block have the same life-cycle and are dispatched onto a GPU streaming multiprocessor at the same time. Depending on the width of the vector functional units/SIMD lanes, threads within a thread-block are grouped into small batches (typically 32 or 64 threads), called a warp or a wavefront. At every cycle, the GPU hardware warp scheduler selects a ready warp from the warp pool for execution (known as the SIMT execution). Warps within a thread-block and threads within a warp are tightly coordinated, bounded by the same synchronization barrier (architecture-specific). To explicitly synchronize threads within a thread-block, {\tt syncthread} can be used. For the particular case of {\tt lud}, the programmer specifies exactly the same number of threads in a thread-block as the warp thread size. Instead of explicitly synchronizing threads with {\tt syncthread}, GEVO finds a kernel variant that leverages the implicit synchronization boundary implemented at the warp granularity and eliminates the {\tt syncthread} call. 

\underline{All}: Depending on the specific implementation of warp and thread-block scheduling policies, additional {\tt syncthread} calls can often be eliminated without altering the execution order between GPU threads ({\tt hotspot}, {\tt lud}, {\tt nw}).

\subsubsection{Removing redundant {\tt store}}
\label{sec:opt_store}

\begin{figure*}[t]
    \centering
    \includegraphics[width=\textwidth]{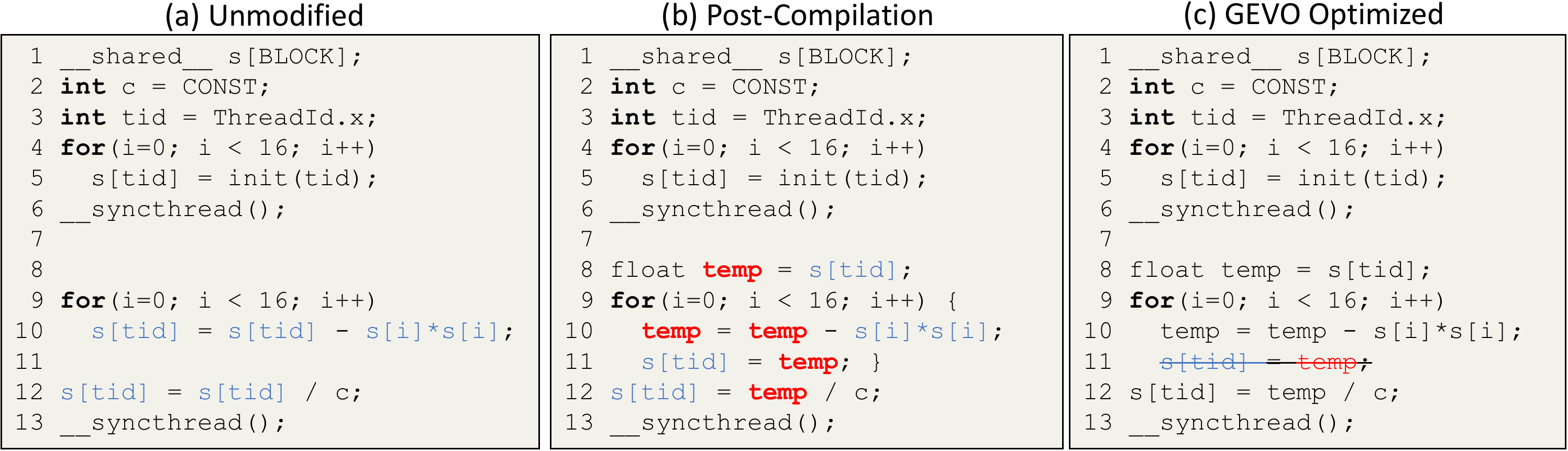}
    \caption{Code snippet from {\tt lud} illustrates a GEVO optimization, which removed redundant {\tt store} instructions.  (a) is the original implementation, (b) is the LLVM/Clang compiled version, and (c) is the optimized kernel variant with GEVO.
    }
    \vspace{-0.2cm}
    \label{fig:lud_store}
\end{figure*}

Another optimized kernel variant discovered by GEVO removed redundant {\tt store} instructions, illustrated with a code snippet from {\tt lud} in Figure~\ref{fig:lud_store}.  We show three versions of the code snippet: (a) is the original implementation, (b) is the LLVM-IR after compilation with the clang compiler, and (c) depicts the optimized kernel variant found by GEVO.
Variable \texttt{s} is a variable stored in the GPU shared memory (Line 1, Figure~\ref{fig:lud_store}(a)) and is initialized in Line 5. Each parallel thread accumulates its own \texttt{s} and reads/writes to \texttt{s} in the shared memory directly (Lines 9-10). Finally, the value of \texttt{s} is written back to the shared memory (Line 12). Compared to updating values in the GPU register file, updating \texttt{s} in the shared memory (Line 10) can incur significant latency and stress the memory subsystem unnecessarily. 

Instead of constantly reading and writing the value of \texttt{s} to the shared memory, the post-compiled LLVM-IR version eliminates the variable reuse patterns for \texttt{s} and replaces \texttt{s} with a temporary variable (\texttt{temp}) which is stored in the register file. To preserve semantic correctness, \texttt{s} in the shared memory is updated with the new value of {\tt temp} (Line 11). Interestingly, this instruction is removed by GEVO. While in theory other threads could access \texttt{s} and receive a stale value, this does not change kernel outputs. GEVO, in this case, trades off program semantics for improved kernel execution time. 
We conjecture that such optimization opportunities, although not completely generalizable, are well-suited for GPUs. This is because GPUs implement more relaxed memory models. General-purpose programs are inherently more sequential and might be less likely to benefit from GPU offloading. If strict ordering between memory operations needs to be enforced, \texttt{threadfence} or \texttt{syncthread} could be inserted. 

\subsection{Application-specific optimizations}
\vspace{-0.1cm}
GEVO discovers optimizations related to the particular application.   We highlight four such optimizations next.
\subsubsection{Removing conditional execution}
\label{sec:opt_if}
GEVO removes dead code that does not affect program output. In the case of GPGPUs, GEVO could eliminate code blocks in the conditional path entirely if the input space does not touch that portion of the kernel. Such kernel variants have been identified for workloads such as {\tt hot}, {\tt lud} and {\tt pf}. 
\subsubsection{Removing redundant load}
\label{sec:opt_load}
\hlnew{
In {\tt bfs}, GEVO removed certain load instructions from a loop which repeatedly loaded data from the same address. In this case, the compiler inserts these load instructions to guarantee the latest updates to the particular memory address will be loaded back correctly in different iterations of the loop. Programmers can avoid these redundant loads if they declare the corresponding variable using a constant modifier, indicating that the variable is read-only for the entire program execution. In this case, GEVO discoverd the data characteristics and addressed the inefficiency without the programmer's involvement. 
}


\subsubsection{Loop perforation}
\label{sec:opt_loop}
Loop perforation is an optimization technique that skips iterations of a loop based on the {\it skip factor} and has been explored for approximate computing~\cite{Sidiroglou:FSE11}. 
GEVO discovers similar optimizations, for example, when loops have been unrolled heavily post-compilation. GEVO then removes some part(s) of the unrolled loop while optimizing the fitness function. We observed this behavior in {\tt sc}, {\tt lud}, and {\tt hot}.
\subsubsection{Memoization}
\label{sec:opt_mem}
Memoization is an optimization technique that stores results of expensive function calls and returns the stored value without re-computation when the same inputs occur again. At the LLVM-IR level, GEVO sometimes identifies similar memoization opportunities by eliminating unneeded instructions and using stored results directly. We find this optimization in the HotSpot temperature modeling tool ({\tt hot}). A kernel in the HotSpot tool performs pre-processing based on the physical dimension of a processor chip. Since the shape of simulated chips is the same across all loop iterations, GEVO discovers memoization opportunities to reuse the preprocessing results of the x-dimension for the y-dimension. \hlnew{Another extreme case was found in {\tt  b+tree}, where one of the input arguments to the program already represents the desired indices in the program output. Thus, GEVO omit almost the entire kernel, leading to more than 5 times performance speedup.}

In summary, we have identified several categories of performance improvements found by GEVO, but we have not studied all such optimizations, and in some cases, we require additional domain-specific knowledge to complete a full analysis. Because GEVO is stochastic, it is not guaranteed to find every possible optimization on every run.

\section{Evaluation of GEVO-mO}
\vspace{-0.1cm}
\label{sec:result_mo}
In this section, we evaluate GEVO (GEVO-mO) in settings where exact output fidelity is not required.  First we consider
GEVO-mO running the Rodinia benchmarks and compare its performance to GEVO-default. We then consider ML workloads and report how GEVO-mo can extract significant performance improvement when domain-specific metrics (model prediction error) are included and co-optimized.

\begin{figure}[t]
\begin{minipage}[t]{0.48\textwidth}
    \includegraphics[width=\textwidth]{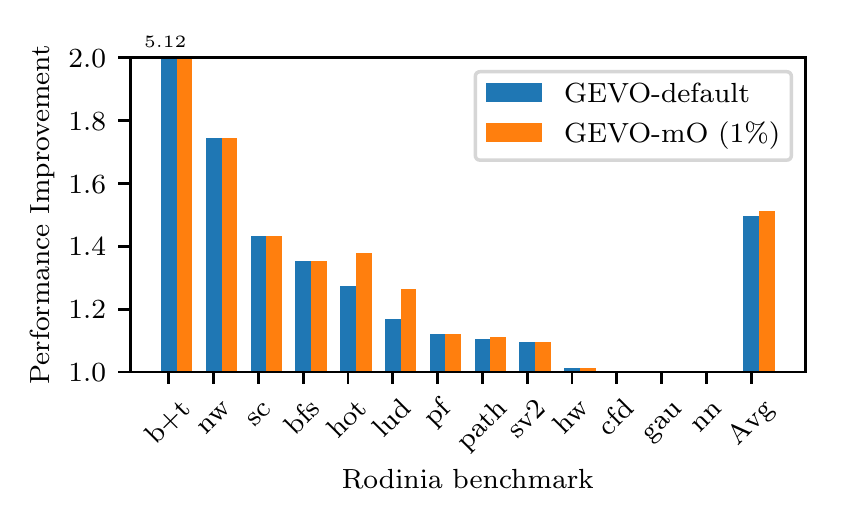}
    \vspace{-0.7cm}
    \caption{Comparison of normalized performance improvement between GEVO-default and GEVO-mO in the Rodinia Benchmark.}
    \label{fig:result_rodinia} 
\end{minipage}\hfill
\begin{minipage}[t]{0.48\textwidth}
    \includegraphics[width=\textwidth]{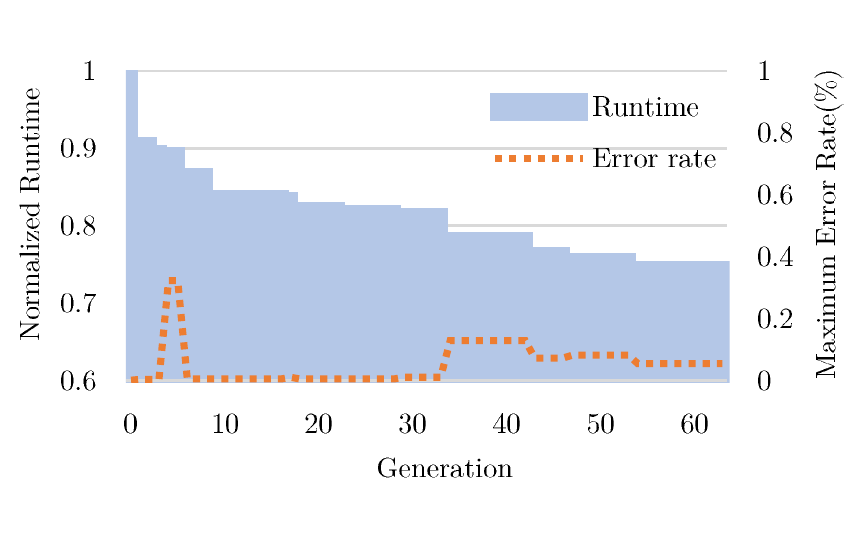}
    \vspace{-0.7cm}
    \caption{Temporal evolution of a {\tt hot} kernel variant.}
    \label{fig:timeline}
\end{minipage}
\end{figure}

\vspace{-0.3cm}
\subsection{Rodinia benchmarks with error tolerance} 
\vspace{-0.1cm}
\label{sec:rodinia_results_mo}

Figure~\ref{fig:result_rodinia} reports overall performance improvement from GEVO-default and GEVO-mO. When accepting up to 1\% kernel output difference, GEVO-mO scavenges additional improvements, reducing run-time by an average of 51.08\% over the baseline. This is mainly achieved by the additional performance improvement in \texttt{hot} and \texttt{lud}, from 27.3\% to 38\% and from 17\% to 26.5\% respectively.

\subsubsection{Temporal Analysis}
To better understand how GEVO-mO co-optimizes runtime and output error, we consider one run of GEVO in closer detail for the program \texttt{hot}, as shown in Figure~\ref{fig:timeline}. On each generation, the figure plots runtime (primary y-axis) and error rate (minor y-axis) for the most fit kernel variant in that generation. As expected, runtime decreases over the run, but the corresponding error rate increases at Generations 5 and 34. This illustrates the design space tradeoff between performance and accuracy. In both cases, GEVO-mO then "repairs" the error rate by introducing other mutations, a phenomenon known as \emph{compensatory evolution}~\cite{Stephan419} in evolutionary biology. 

There are three key mutations in the last generation. When combined, they reduce the error rate to less than 0.1\%, whereas if individually applied, the error rate is much higher at 0.3\%. This highlights the strength of a population-based search method like GEVO---sub-optimal individuals in one generation can be combined and/or serve as a stepping stones to the discovery of successful combinations of mutations. Further, the best kernel variant would not be found if a tighter error bound had been enforced from the beginning.

\subsubsection{Optimization analysis}
\label{sec:opt_mo_rodinia}
For \texttt{hot} and \texttt{lud}, mutation analysis reveals how additional improvements are achieved.

\underline{\tt \bf hot}: Additional performance improvement is achieved by removing additional synchronization primitives. This optimization raises the possibility of a race condition, because the outputs vary slightly from run to run but always remain under the 1\% threshold. Although race conditions are a potential concern, earlier work proposes lock-free approaches for specific algorithms and includes a proof that the algorithm can converge even with a race condition~\cite{recht2011hogwild}.

Our analysis of the \texttt{hot} optimization provided two possible explanations for how removing the synchronizations leaves a viable program.  First, in \texttt{hot}, each thread updates the temperature of a spot in a 2-dimension grid, based on the ambient temperature and the temperature of the surrounding spots. Synchronization on each time step allows every other thread to calculate and update its temperature based on up-to-date temperature values nearby. However, there are two levels of synchronization in \texttt{hot}: in-kernel synchronization using the \texttt{syncthreads} function and a global synchronization when the kernel is relaunched. Removing the \texttt{syncthreads} call inside the kernel implies that the temperature over the grid will be synchronized only every other time step, which might leave the simulation within its error tolerance. 
The second explanation derives from the simulation setting. In \texttt{hot}, the temperature difference between the ambient and initial grid temperature (ambient is set to 80\textdegree K while the grid is mostly between 320\textdegree K and 345\textdegree K) is much larger than the difference between a given spot and its neighbors (usually within 10\textdegree K).  As a result, the ambient temperature contributes more than surrounding temperature to the thermal simulation. This effect mitigates the effect of the inaccuracy introduced by removing the synchronization.

\underline{\tt \bf lud}: 
GEVO finds performance improvement by reusing the result from an earlier iteration of a loop and avoiding computation in the latter iteration because the loop has been unrolled by the compiler. 
This optimization is an example of memoization introduced in the previous section. \texttt{lud}, standing for Lower-Upper decomposition decomposes a given matrix into a product of two triangular matrices where each one has the lower/upper part of matrix filled with zeroes. Although it might seem unacceptable to tolerate any error in the solution of a linear system, a method known as incomplete LU factorization~\cite{meijerink1977iterative} approximates the solution of LU decomposition for lower computational cost. We suspect that GEVO accidentally re-discovered this technique for improving \texttt{lud}'s performance, providing a nice example of approximate computing.

\begin{figure*}[t]
    \includegraphics[width=1\textwidth]{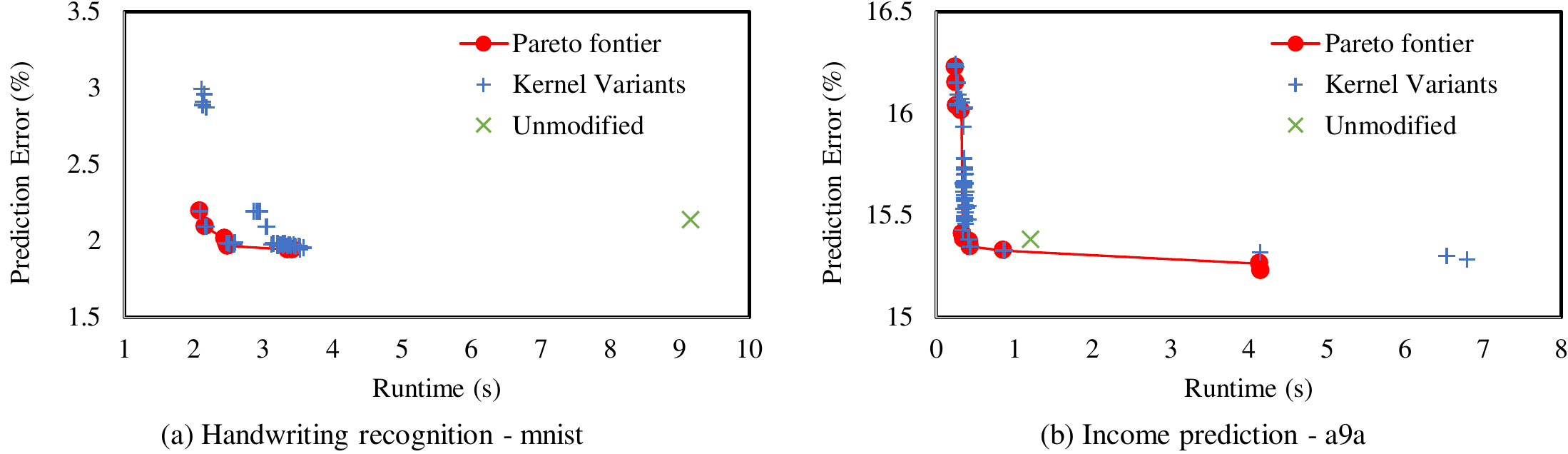}
    \vspace{-0.4cm}
	\caption{
	Pareto-frontiers and other kernel variant measurements for (a) the handwriting recognition (SVM with MNIST) and (b) the income prediction (SVM with a9a).}
	\label{fig:thundersvm_pf} 
\end{figure*}

In other Rodinia benchmarks, GEVO failed to find significant improvements when output fidelity was relaxed. Although we don't know why these applications were more challenging, there are several possibilities, including the most obvious one that the 1\% error tolerance based only on the raw kernel output from the GPU is too constrained to allow appreciable expansion of the optimization search space. 
Therefore, in the next section, 
we change the error definition from GPU kernel output difference to application-defined error.
For ML, the natural application error is model prediction error. This change allows GEVO to find application-specific optimizations that produce significantly different kernel output while maintaining model accuracy. 

\subsection{Evaluation of Machine Learning Applications (Dataset-specific optimizations)}
\vspace{-0.1cm}
\label{sec:ml_results}

Machine learning (ML) is a popular class of intrinsically error-tolerant applications, which consume large computational resources, and is particularly suitable for the GEVO approach. We consider two ML models, SVM and ResNet18, and use them to illustrate how performance and accuracy can be co-optimized. Although earlier work 
examined accuracy/runtime tradeoffs of implementations~\cite{huang2017speed}, we are unaware of earlier work targeting genetic improvement of ML LLVM-IR kernels.

\subsubsection{SVM}
\label{sec:svm_results}

Figure~\ref{fig:thundersvm_pf} shows the Pareto frontiers found by GEVO for handwriting recognition (SVM with {\tt MNIST}) and income prediction (SVM with {\tt a9a}). The x-axis represents the measured kernel runtime and the y-axis represents the training inference prediction error in \%.
We report results for each kernel variant in GEVO's final generation, relative to the original unmodified kernel. 
Figure~\ref{fig:thundersvm_pf} also shows how GEVO-mO navigates away from the original, sub-optimal, kernel implementation and explores the better performing part of the search space.

\begin{wrapfigure}{R}{0.5\textwidth}
\lstset{
  style=2column,
  moredelim=[is][\sout]{|}{|},
  moredelim=[is][\bfseries\color{red}]{<<<}{>>>},
}
\begin{lstlisting}
...
while (1)
  // select f Up
  if (is_I_up(...))
    f_val_reduce[tid] = f;
  up_val = f_val_reduce[...];
  
  // select f Low
  if (is_I_low(...))
    // f_val_reduce[tid] = -f;
    <<<f_val_reduce[tid] = 1 - f;>>>
  down_val = f_val_reduce[...];
  
  if (up_val - down_val < epsilon)
    break;
\end{lstlisting}
\vspace{-0.5cm}
\caption{Code snippet from ThunderSVM illustrates an optimization discovered by GEVO. The comment at line 10 is the original code and line 11 indicates the GEVO modification. }
\label{fig:svm_mutate}
\end{wrapfigure}

Considering the kernel variant in the Pareto frontier that represents the best combined improvement,
we find that GEVO-mO achieves 3.24X improvement for handwriting recognition (MNIST) and 2.93X performance improvement for income prediction (a9a)'s kernel performance, which increases overall model training speed by 50\% and 24.8\% respectively.  At the same time, accuracy on the training set improved from 97.86\% to 98.03\% (MNIST) and from 84.61\% to 84.65\% (a9a), which was unexpected as we imagined the optimization would tradeoff accuracy against training time. Next, we tested the trained GEVO-optimized models on their official testing datasets, where we find accuracy is improved slightly,
from 98.37\% to 98.5\% (MNIST) and from 84.59\% to 84.64\% (a9a).

We also consider whether the SVM evolved for training a specific dataset can achieve similar improvements on a different dataset in the same class (after all, that would be the main advantage of optimizing the training procedure for a particular type of application). We tested SVM optimized for the MNIST common dataset (60,000 samples) by using it to train the large MNIST dataset (8,000,000 handwriting samples). Since the large MNIST dataset does not have a separate testing dataset, we report the 10-fold cross validation accuracy for the model from unmodified and optimized ThunderSVM, which are 100\% and 99.997\% with the respective training time in 1182 and 121 minutes. 

Our optimization analysis, shown in Figure~\ref{fig:svm_mutate}, shows how GEVO changes the termination condition of a while loop, by increasing the lower bound by 1 in line 11. As a result, there is a chance of producing a smaller value in the \texttt{if} statement in line 14, causing the execution to exit the while loop sooner.
This loop implements a SVM solver using \emph{sequential minimal optimization}, which iteratively approaches the optimal solution, terminating when progress has slowed. 
Thus, GEVO relaxes the convergence condition, which would normally be expected to reduce solution correctness. However, for MINST, this change actually improves model accuracy, perhaps by avoiding overfitting. 
We leave further analysis of this surprising result for future work.

\subsubsection{ResNet18}
\label{sec:caffe2_results}

\begin{figure*}[t]
    \includegraphics[width=1\textwidth]{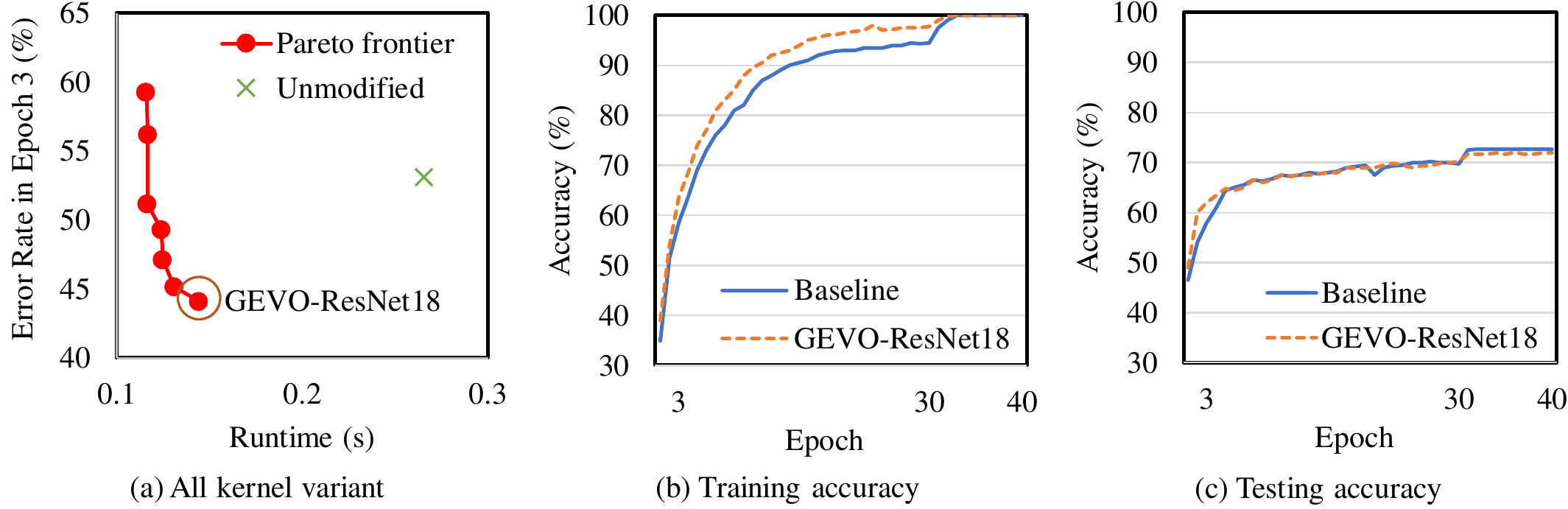}
	\caption{
	(a) Pareto-frontiers for the image classification (ResNet18 with CIFAR-10). The kernel in the circle is manually selected to retrain the model until model converges, with its (b) training accuracy and (c) testing accuracy across epochs.}
	\label{fig:caffe2_train_test} 
\end{figure*}

Similar to SVM, figure~\ref{fig:caffe2_train_test}(a) presents the Pareto-frontier for image classification (ResNet18 with CIFAR-10). Here we select the kernel variant that gives the best training accuracy on the third epoch (56\% compared to 47\% in the Baseline).  This variant is also 1.79X faster than original kernel. However, the kernel contributes less than 1\% of the entire training time.  Recall that we do not have access to the other kernels for ResNet18 operators (Section~\ref{sec:benchmarks}). 

Recalling that we train ResNet18 for only three epochs during GEVO's optimization searches and only on the momentumSGD kernel, we first examine the kernel performance throughout the training process until the point where model accuracy has converged. Figures~\ref{fig:caffe2_train_test}(b) and~\ref{fig:caffe2_train_test}(c) report training and testing accuracy across epochs for the original baseline kernel and a GEVO-optimized variant. Considering training accuracy, the GEVO-optimized kernel consistently beats the baseline across epochs by up to 7\% until epoch 30, when both kernels achieve 100\% training accuracy. The sudden jump in training accuracy around epoch 30 is caused by a learning rate change in epoch 29, which occurs in code that is not available to GEVO. If we ignore that external learning rate change, the GEVO-optimized kernel would converge at epoch 25, and the baseline would converge at epoch 28, with 2\% lower training accuracy. Turning to testing accuracy, both kernels have comparable accuracy through epoch 30 when the learning rate is changed. After epoch 30, the baseline is 1\% more accurate than the variant (73.24\% to 72.31\%). 

\begin{wrapfigure}{R}{0.5\textwidth}
\lstset{
  style=2column,
  moredelim=[is][\sout]{|}{|},
  moredelim=[is][\bfseries\color{red}]{<<<}{>>>},
}
\begin{lstlisting}
/*    N = number of parameter
 * m[i] = momentum
 * g[i] = gradient
 * BETA = momentum decay rate
 *   LR = learning rate
 */
for (i=tid; i<N; i+= |GRID_SIZE| <<<N>>>) {
  float mi = m[i];
  float mi_new = BETA*mi + LR*g[i];
  m[i] = |mi_new| <<<LR*g[i]>>>;
  g[i] = (1+BETA)*mi_new - BETA*mi;
  
  if (param) 
    param[i] -= g[i];
}
\end{lstlisting}
\vspace{-0.5cm}
\caption{Code snippet from Caffe2 momentumSGD operator illustrates two optimization discovered by GEVO. }
\label{fig:caffe2_mutate} 
\end{wrapfigure}

Figure~\ref{fig:caffe2_mutate} shows an important code optimization found by GEVO in the Caffe2 momentumSGD operator. There are two modifications leading to the accuracy and performance improvements. First, GEVO changes the loop boundary so the loop is executed only once (line 7 in Figure~\ref{fig:caffe2_mutate}). The momentumSGD kernel updates the parameters (weight and bias) of the neural network and the loop here represents how many parameters need to be updated.  Thus, GEVO's success when reducing the number of loop iterations suggests that the ResNet18 model is overly complicated for the CIFAR-10 dataset. This is similar to weight pruning~\cite{Molchanov2016PruningCN} or hyperparameter search~\cite{bergstra2012random} for a particular dataset. 
GEVO also changes how the momentum is calculated by using only the current gradient and not considering the prior gradient.  This optimization illustrates how GEVO can tailor an algorithm to a particular dataset.


%% file: discussion.tex
\section{Discussion}
\label{sec:discussion}
\vspace{-0.1cm}

This paper presents GEVO, a new method that uses stochastic population-based search to discover optimizations of GPGPU kernels.
GEVO trades off absolute program semantics for other important non-functional design aspects. The experimental results demonstrate that by relaxing program semantics, GEVO can find novel and substantial improvements, both for runtime alone, and for the case of multiple optimization objectives, e.g., accuracy and runtime. The proposed approach, while not intended for applications with critical correctness requirements (e.g., inner loops of avionics software or some systems programs), is suitable for many other applications, including the important class of ML codes. When we consider handwriting recognition, income prediction, and image recognition ML workloads, our results show that GEVO explores the optimization search space and finds multiple points along the Pareto frontier that maximize trade offs between performance and accuracy. In some cases, however, GEVO can ``have the best of both worlds" by finding a significant 3.24x speedup of the handwriting recognition kernel (SVM with {\tt MNIST} and achieve modest improvements of prediction accuracy. This translates to 50\% training time reduction with 0.17\% improvement on the prediction accuracy, reflecting absolute improvements in both dimensions.  

By design, GEVO does not aspire to preserve exact program semantics, and in many cases, it can identify algorithmic improvements that are inaccessible to methods that require complete semantic consistency with the original program.  In other circumstances, however, GEVO could potentially find optimizations that break required semantic properties which are not enforced by the test suite.  Optimizations that
relax memory synchronization requirements provide a good example. In the cases we examined, eliminating redundant synchronizations did not affect program behavior. However, this strategy is 
risky in general because it depends on specific memory access patterns, which in turn rely on the combined effect of the target application and its execution environment (hardware architecture and system software).  There is currently a great deal of research interest in studying how memory accessing order can be relaxed for better performance. This includes application-specific approaches, e.g., to schedule and prioritize memory access for specific tasks~\cite{nguyen2013lightweight, alistarh2015spraylist} and system-level approaches such as non-blocking or wait-free synchronization with system or architecture support~\cite{xiao2010inter, chung2010asf}.  GEVO could potentially be applied to identify these optimization spots for researchers to further analyze when necessary. 

More generally, as we learn more about when and how GEVO succeeds and fails, we foresee new methods for post-hoc validation of evolved codes, e.g., by synthesizing new test cases on the fly to test synchronization, or ultimately, using
program analysis methods to highlight semantic differences between original
and evolved kernels~\cite{padACE}.

\hlnew{We have also explored the idea of enhancing the proposed design by considering system-specific architecture features. Cache efficiencies concern the performance of GPUs. Thus, we extended the scope of GEVO to explore the performance optimization space of the GPU cache configuration. In this case, we enabled GEVO to control cache bypassing by introducing {\tt ld.cg} (for caching at the L2 cache but bypassing the L1 cache) and {\tt ld.ca} (for caching at both the L1 and L2 caches) to the genetic operations. By doing so, the mutation operation can specify whether data are bypassed from the GPU L1 cache at the granularity of instructions in the NVIDIA PTX ISA (By injecting inline-assembly in LLVM-IR) and discover performance speedup opportunities, similar to on-demand cache fetching~\cite{jia2012demandfetchcache} or cache bypassing optimization~\cite{lee2016ctrl, arunkumar2016id}. Overall, this mutation operator did not produce frequent enough
performance improvements to justify adding it to GEVO's mutation suite.
 It seems that GEVO often finds equivalent optimizations without
explicitly using the cache-specific mutation, simply by
moving load instructions.}


The results reported here are specific to the programs, inputs, and the particular GEVO runs we studied. There were some programs for which GEVO was unable to find improvement. Thus, further experimentation is required to understand the generality of our results.
GEVO found application-specific, architecture-specific, and dataset-specific optimizations. In future work, we plan to test GEVO on other applications and analyze more carefully why some programs admit significant improvements and others do not. 
Since GEVO's approach is agnostic about optimization criteria, it is easy to imagine other compelling optimizations. For example, GEVO could customize the LLVM-IR for particular classes of inputs, or even generate diverse versions of the kernel, each of which uses a different power budget, to defeat some power side channel attacks. 

GEVO itself has many possible parameter settings, including population size, mutation and crossover rates, and there are many existing evolutionary algorithms with different strategies for selection and multi-objective function optimization.  We began with the most popular multi-objective framework (DEAP), modified it for our application, and conducted several initial experiments to find a configuration that works well for GPGPU optimization.  However, it is certainly possible that other evolutionary algorithms or other parameter settings for GEVO would produce better results. Similarly, we chose a 1\% error tolerance arbitrarily for the GEVO-mO experiments.  Acceptable errors may vary across programs, and in some cases, could translate into improved optimizations. \hlnew{An example can be found in Yazdanbakhsh's work~\cite{yazdanbakhsh2016axbench} where {\tt srad} from the Rodinia benchmark could accept up to 10\% error, with additional optimizations being revealed. Increasing GEVO's error tolerance to 10\% for applications such as this could lead to additional optimizations.}

Our
mutation operators are more expensive than those used in earlier work on genetic improvement of software.  The additional cost arises from the nature of the single static assignment discipline in the LLVM-IR.  As a result, as Table~\ref{table:benchmarks} shows,
GEVO searches for a very low number of generations on many of the benchmarks.  A general rule-of-thumb would suggest running the EC search for at least as many generations as there are individuals in the population (250 in our experimental setup). With additional computational resources, we could expect much better performance improvements, especially on the benchmarks that ran for fewer than thirty generations.  

It is well-known that significant human expertise is required to tune a ML model to extract the best performance on a particular task. For instance, in SVM, the regularization parameter known as $C$ is manually determined to balance the generalization and the training error (underfitting versus overfitting).  Similarly, in neural networks human expertise and experimentation is used to find an appropriate network architecture, which determines how many neurons are used and how they are connected.
Currently, these design decisions are determined empirically for each dataset, and often tested repeatedly. Automating these design decisions, commonly referred as \textit{hyperparameter search}~\cite{bergstra2012random} or \textit{AutoML}~\cite{Thornton:2013:ACS:2487575.2487629}, 
to reduce human effort is a current research direction, and many algorithms have been proposed, including simple grid search~\cite{Larochelle:2007:EED:1273496.1273556}, random search~\cite{bergstra2012random}, reinforcement learning~\cite{zoph2016neural, bello2016neural}, evolutionary computation~\cite{neat}, and gradient decent~\cite{liu2018darts}.  These hyperparameter search algorithms differ from GEVO in that they do not touch the underlying code implementation.  The results presented here show that GEVO can discover effects that are similar to those found with hyperparameter search, at the same time that it improves the implementation itself. 
\hlnew{This suggests that GEVO is capable of performing hyperparameter search along with network optimizations. We plan to conduct additional experiments expanding the work described here, with the MLPerf Training and Inference Benchmark Suites~\cite{mattson2020mlperf,mlperf-training,mlperf-inference}.}


\section{Conclusion}
\vspace{-0.1cm}
\label{sec:conclusion}
Programmers today develop software that is expected to meet functional correctness, to avoid opening up new security vulnerabilities, and to execute efficiently.  The software is often developed on complex programming stacks, is compiled to run on complex proprietary architectures which lack transparency, and it often has unanticipated interactions with runtime environments and workloads.
This paper presents one approach to managing this complexity, focusing on code optimization for GPUs.  GPUs are an appealing target because they are widely used to accelerate compute-intensive applications and because GPU codes are often error-tolerant and amenable to approximate optimization methods that do not guarantee exact semantic equivalence to the original program.

Our approach, implemented as GEVO, uses population-based stochastic search on LLVM-IR GPU programs. GEVO finds versions of programs that retain required functionality, as assessed by test cases, and optimize one or more fitness criteria.  We focus on the dual objectives of minimizing runtime and application error, finding significant reductions in runtime for most programs with little or no penalty in output error.  GEVO finds optimizations that leverage details about the architecture, the application design, and even particular workloads.  For large-scale computation-intensive applications such as ML, we hope that the methods presented here can contribute to improved software deployments.  We also hope that some of the unusual optimization opportunities identified by GEVO will lead to improved software development practices, whether through improved tools or through improved awareness on the part of application developers.

\vspace{-0.1cm}